\documentclass[lettersize,journal]{IEEEtran}
\usepackage{amsmath,amsfonts}
\usepackage{algorithmic}
\usepackage{algorithm}
\usepackage{array}
\usepackage[caption=false,font=normalsize,labelfont=sf,textfont=sf]{subfig}
\usepackage{textcomp}
\usepackage{stfloats}
\usepackage{url}
\usepackage{verbatim}
\usepackage{graphicx}
\usepackage{cite}
\usepackage{color}
\usepackage{amsfonts}
\usepackage{amssymb}
\usepackage{booktabs}
\usepackage{multirow}
\usepackage{threeparttable}
\usepackage{makecell}
\usepackage{bbding}
\usepackage[switch]{lineno}

\newcommand{\ie}{\textit{i}.\textit{e}.}

\usepackage{hyperref} % 参考文献、引用章节、图片和表格跳转
\hypersetup{hidelinks}

\begin{document}
% \linenumbers
\title{RCNet: Deep Recurrent Collaborative Network for Multi-View Low-Light Image Enhancement}
% Multiple Views in the Dark: Recurrent Refinement Learning with Cross-View Attention for Multi-View Low-Light Image Enhancement

%\author{IEEE Publication Technology,~\IEEEmembership{Staff,~IEEE}
\author{Hao Luo, Baoliang Chen,~\IEEEmembership{Member,~IEEE}, Lingyu Zhu,~\IEEEmembership{Student Member,~IEEE}, Peilin Chen and Shiqi Wang,~\IEEEmembership{Senior Member,~IEEE} % <-this % stops a space

\thanks{This work was supported in part by ITF Project GHP/044/21SZ, in part by RGC General Research Fund 11203220/11200323, and in part by the National Natural Science Foundation of China under Grant 62401214. (\textit{Corresponding author: Shiqi Wang.})}
\thanks{Hao Luo, Lingyu Zhu, Peilin Chen, and Shiqi Wang are with the Department of Computer Science, City University of Hong Kong, Hong Kong (e-mail: hluo29-c@my.cityu.edu.hk; lingyzhu-c@my.cityu.edu.hk; plchen3@cityu.edu.hk; shiqwang@cityu.edu.hk).}
\thanks{Baoliang Chen is with the Department of Computer Science, South China Normal University, China (e-mail: blchen@scnu.edu.cn).}% <-this % stops a space
}
%\thanks{Manuscript received April 19, 2021; revised August 16, 2021.}}

\markboth{Journal of \LaTeX\ Class Files,~Vol.~14, No.~8, August~2021}%
{Shell \MakeLowercase{\textit{\textit{et al.}}}: A Sample Article Using IEEEtran.cls for IEEE Journals}


\maketitle

\begin{abstract}
Scene observation from multiple perspectives would bring a more comprehensive visual experience. However, in the context of acquiring multiple views in the dark, the highly correlated views are seriously alienated, making it challenging to improve scene understanding with auxiliary views. Recent single image-based enhancement methods may not be able to provide consistently desirable restoration performance for all views due to the ignorance of potential feature correspondence among different views. To alleviate this issue, we make the first attempt to investigate multi-view low-light image enhancement. First, we construct a new dataset called Multi-View Low-light Triplets (MVLT), including 1,860 pairs of triple images with large illumination ranges and wide noise distribution. Each triplet is equipped with three different viewpoints towards the same scene. Second, we propose a deep multi-view enhancement framework based on the Recurrent Collaborative Network (RCNet). Specifically, in order to benefit from similar texture correspondence across different views, we design the recurrent feature enhancement, alignment and fusion (ReEAF) module, in which intra-view feature enhancement (Intra-view EN) followed by inter-view feature alignment and fusion (Inter-view AF) is performed to model the intra-view and inter-view feature propagation sequentially via multi-view collaboration. In addition, two different modules from enhancement to alignment (E2A) and from alignment to enhancement (A2E) are developed to enable the interactions between Intra-view EN and Inter-view AF, which explicitly utilize attentive feature weighting and sampling for enhancement and alignment, respectively. Experimental results demonstrate that our RCNet significantly outperforms other state-of-the-art methods. All of our dataset, code, and model will be available at https://github.com/hluo29/RCNet.
\end{abstract}

\begin{IEEEkeywords}
Multi-view low-light enhancement, collaborative network, intra-view enhancement, inter-view alignment \& fusion.
\end{IEEEkeywords}

\section{Introduction}
% {介绍low-light任务的重要性和必要性，以及low-light enhancement这个task存在的困难}
% {暗光下不同视角亮度和噪声分布不同的客观存在性}
\IEEEPARstart{W}{hen} capturing images from different viewpoints in the dark, the imaging process of each view would suffer from certain degrees of quality degradation, e.g., insufficient illumination and intensive noise. As a result, the low-light images not only attenuate the human visual perception intuitively, but also pose grand challenges to outdoor recognition tasks, such as object detection \cite{ren2015faster, lin2017focal} and semantic segmentation \cite{zhao2017pyramid, chen2018encoder}. For a single object in the 3D world, there often appear diverse reflectances due to the uncertainty of illumination intensity as well as noise distribution from different viewpoints \cite{oxholm2014multiview}. To some extent, this capricious situation suggests the collaborative restoration of similar regions across different views in the same scene, which has been unfortunately ignored by recent single image-based low-light enhancement methods. In this paper, we focus on a new research problem \textit{multi-view low-light image enhancement} by building a new multi-view dataset and developing a novel algorithm with the philosophy of collaborative enhancement.

\begin{figure}[t]
\centering
\includegraphics[width=\linewidth]{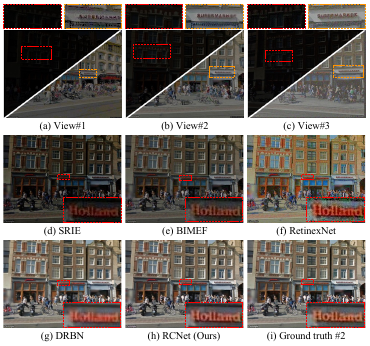}
\caption{Illustration of multi-view low-light images and the enhanced results of state-of-the-art methods. (a)$\sim$(c): three different views in the same scene, with each composed of low-light image and bright result corrected by Gamma transformation. (d)$\sim$(h): the results of SRIE\cite{fu2016weighted}, BIMEF\cite{ying2017bio}, RetinexNet\cite{wei2018deep}, DRBN\cite{yang2020fidelity} and our RCNet, using the low-light View\#2 as input. (i): the normal-light version of low-light View\#2.}
\label{fig1-mvll}
\end{figure}
% The first row visualizes the enlargements of selected regions in the second row (a)-(c), and the second row shows three different views in the same scene, with each composed of low-light image and the bright result corrected by Gamma transformation. The third (d)-(f) and fourth (g)-(i) rows example the results of SRIE\cite{fu2016weighted}, BIMEF\cite{ying2017bio}, RetinexNet\cite{wei2018deep}, DRBN\cite{yang2020fidelity}, and our proposed RCNet with the low-light View\#2 as input.

% {当前low-light的文章进行大致分类，并指出现方案的可能不足，说明单幅图像增强任务的non-trivial，引出需要更多的额外信息帮助恢复}
% {大范围概括<传统方法>，包括HE和Retinex，结尾指出handcraft设计的通病}
Traditional low-light image enhancement methods attempt to recover normal-light images using histogram equalization (HE) \cite{yousuf2011effective, deng2010generalized} by stretching the dynamic range of dark image directly, or utilizing decomposition-based Retinex theory \cite{rahman2004retinex, li2018structure, fu2015probabilistic} by assuming a dark image as the combination of reflectance and illumination components. Moreover, some methods \cite{ying2017new, ren2018lecarm} focus on the response properties of cameras to recover low-light images by the estimations of the camera response model and exposure ratio map. However, these methods are specifically designed through handcrafted priors or models and are easily accompanied by a series of image artifacts, e.g., noise amplification and lightness distortion.

% {大范围概括<Deep方法>，结尾点出只适用于Single image/Single view}
With the prevalence of convolutional neural network (CNN) \cite{hidayati2018dress, cheng2021fashion}, deep learning begins to be introduced into low-light enhancement and achieves significant quality improvements. Wei \textit{et al.} \cite{wei2018deep} first proposed to combine the Retinex theory with CNN, in which cascaded convolutional layers were developed to predict the decomposed reflectance and illumination. Following \cite{wei2018deep}, many advanced methods have been proposed by either exploring more efficient combination modes between Retinex decomposition and CNN \cite{zhang2019kindling, liu2021retinex, zhao2021retinexdip}, or instead designing a fully CNN-based enhancement architecture \cite{lv2018mbllen, lim2020dslr, wang2020lightening, zamir2020learning, yang2021band}. However, all above methods are mainly applicable to single image low-light enhancement (\textit{i.e.}, single view in one scene) and are prone to neglect the strongly correlated correspondence between different dark views when directly applied to multi-view low-light vision (\textit{i.e.}, multiple views in one scene). This may cause color distortion or blurry texture in the enhanced images, as illustrated in Fig.~\ref{fig1-mvll}. 
%, as demonstrated in experimental results.

% {结合Fig.1重点分析<多视角暗光>}
However, this ill-posed enhancement problem can be greatly alleviated via multi-view collaboration, which aims to search the most similar textures across neighboring views. In general, different low-light views even in the same scene have extremely different degrees of degradation. As shown in Fig.~\ref{fig1-mvll} (a)$\sim$(c), two important findings could be observed regarding multi-view low-light imaging: (1) \textit{for different viewpoints, the same object usually presents various degrees of visibility}. For example, some windows of the building in View\#2 and View\#3 are easier to observe than those in View\#1, as highlighted in red dotted box; (2) \textit{the difference of noise distribution across multiple views contributes to noise suppression by similar regions from auxiliary views}. In order to explore whether the noise distribution also differs across diverse views, we adopt the Gamma transformation to adjust the lightness of low-light images. As shown in the yellow dotted box, compared to letters in View\#1 and View\#3, those in View\#2 tend to appear smoother. In short, these two findings imply the significance of multi-view collaboration (via the comparison from Fig.~\ref{fig1-mvll} (d)$\sim$(i)) and motivate us to investigate the multi-view low-light image enhancement.

% {概述提出的解决方案，并分<四点>进行总结}
In this paper, we first construct a new dataset called Multi-View Low-light Triplets (MVLT), including 1,860 pairs of triple images with large illumination variations and random noise distribution. Each triplet is equipped with three different viewpoints towards the same scene. Then we propose a deep multi-view enhancement framework based on Recurrent Collaborative Network (RCNet). In contrast to single image-based enhancement methods which ignore the potential feature correspondence among different views, our method achieves multi-view low-light image enhancement in recurrent view collaboration. The intra-view feature enhancement followed by inter-view feature alignment and fusion is performed to model the intra-view and inter-view feature propagation sequentially. In this way, the enhanced result would benefit from auxiliary views with effective lightness correction and noise suppression. Besides, our network can efficiently cope with the large changes of viewpoints in recurrent steps. Experimental results demonstrate that our RCNet significantly outperforms other state-of-the-art methods.

In summary, the main contributions of this paper are listed as follows,
{
\begin{itemize}
\item We build a large-scale multi-view low-light dataset with a total of 1,860 pairs of low- and normal-light images, \textit{i.e.}, 620 triples of multi-view low-light pairs. This dataset provides diverse multi-view scenes with various illuminant ranges as well as random noise distribution.

\item We propose a novel multi-view enhancement framework RCNet, in which intra-view feature enhancement followed by inter-view feature alignment and fusion is designed to benefit from similar feature correspondence across different views. 

\item We further develop two different modules E2A and A2E to enable the interactions between Intra-view EN and Inter-view AF, enabling attentive feature weighting and sampling for enhancement and alignment, respectively.

%\item We conduct various experiments to validate the effectiveness of our RCNet. The quantitative and qualitative results demonstrate that our method achieves the state-of-the-art performance on quality fidelity and perception. 
\end{itemize}
}

% {文章框架分支概述}
%The rest of this paper is organized as follows. In Section \ref{sec2-relatedwork}, we review the related traditional and learning-based low-light image enhancement methods. In Section \ref{sec3-dataset}, we introduce the synthesis procedure and statistical characteristics of the built MVLT dataset. Section \ref{sec4-method} presents the detailed network framework for multi-view low-light enhancement. Experimental results and concluding remarks are discussed in Section \ref{sec5-experiments} and Section \ref{sec6-conclusion}, respectively.

\section{Related Work} \label{sec2-relatedwork}
\subsection{Traditional Low-light Enhancement}
In order to mitigate low-intensity pixel values with narrow distribution in low-light images, histogram equalization (HE) is often used to stretch out the illumination range for contrast enhancement. In the early stage, Global-based HE \cite{yousuf2011effective, castleman1996digital} usually adopted the entire low-light image histogram statistics as the mapping function to improve image contrast, but cannot adapt with local illumination information. To resolve this problem, local-based HE \cite{deng2010generalized, kim2001advanced} performed repeated sub-block histogram equalization within the sliding window, making full use of the local brightness features. Essentially speaking, the overlapped sub-block equalization methods have to take large computational costs and much time to find a well-performed block size for noise suppression. Therefore, several HE-based methods were proposed to achieve efficient contrast improvement. Abdullah \textit{et al.} \cite{abdullah2007dynamic} designed a Dynamic HE to deal with biased transformation via partitioning operation. Each sub-histogram with a controlled dynamic range can avoid losing histogram components and preserve the details in the enhanced result. In \cite{kim1997contrast, wang1999image}, the dark image histogram was divided into two different parts using preset illumination values to preserve the original mean brightness in the resultant image. However, when this two-part division was extended into exponential times via recursive sub-histogram equalization \cite{chen2003contrast}, the enhanced result is almost the same with input degraded image in low-light enhancement.

Inspired by the retina-and-cortex system of human vision, the Retinex theory \cite{land1977retinex} is applied for low-light enhancement, which defines the dark image as the combination of reflectance and illumination components. Several multi-scale variants \cite{rahman2004retinex, jobson1997multiscale} of Retinex have been designed to improve the generalization towards diverse images. Lee \textit{et al.} \cite{lee2013adaptive} developed an adaptive weight between each single-scale Retinex and the dark input, to enhance the naturalness and color rendition in every region of the image. More efforts have also been made to reflectance and illumination estimation \cite{guo2016lime, fu2015probabilistic, fu2016weighted, li2018structure, ren2018joint}. These methods show impressive enhancement performance with specially hand-crafted constraints, which may be hardly applied to those low-light images with complex noise distribution and large illumination changes.

\subsection{Deep Learning-based Low-light Enhancement}
In \cite{shen2017msr, lv2018mbllen}, learning-based neural networks began to be introduced to restore low-light images and achieved significant performance improvement. Later, Li \textit{et al.} \cite{li2020luminance} optimized the low-light enhancement network in a coarse-to-fine strategy \cite{makwana2024livenet}, including coarse contrast feature extraction and luminance-aware pyramid refinement. Instead of learning direct mapping from low-light image to normal-light counterpart, numerous efforts have been dedicated to residual learning \cite{lim2020dslr, wang2020lightening, zamir2020learning, yang2020fidelity, yang2021band}, frequency decomposition \cite{xu2020learning, zhu2022enlightening}, degradation decoupling \cite{jiang2022degrade, guo2023low}, and guided fusion \cite{lu2020tbefn, xu2022snr}. In \cite{lim2020dslr, zamir2020learning}, a multi-scale residual block was frequently adopted to propagate spatially-precise high-order features \cite{hsieh2019fashionon, li2020mucan}, and the enhancement result can be obtained by a learned residual. Inspired by the low-light color image formulation, Jiang \textit{et al.} \cite{jiang2022degrade} designed a degradation-to-refinement generative network to estimate the environment illumination color distortion followed by the diffuse illumination color refinement. Guo \textit{et al.} \cite{guo2023low} proposed to decouple the entanglement of noise and color distortion by performing noise removal and color correction along with illumination adjustment. Similar attempts could also be observed in other image-based tasks, such as rain streaks decomposition in rain removal \cite{jiang2020decomposition, jiang2021rain}, transmission maps decomposition in image dehazing \cite{yang2022self}, and the facial action units \cite{xie2022overview, xie2020assisted, lo2021facial}. In \cite{xu2022snr}, Xu \textit{et al.} estimated the signal-of-noise-ratio map to guide the combination between long-range and short-range features for spatial-varying enhancement.

Recent works also integrated the Retinex theory into deep networks \cite{wei2018deep, zhang2019kindling, liu2021retinex, zhao2021retinexdip, yi2023diff}. Wei \textit{et al.} \cite{wei2018deep} first built the RetinexNet with three modules including decomposition, adjustment, and reconstruction. In \cite{yi2023diff}, Yi \textit{et al.} decoupled the low-light image enhancement into Retinex decomposition and conditional image generation to utilize the advantages of physical model and generative network, respectively. In addition, Jiang \textit{et al.} \cite{jiang2021enlightengan} proposed a global-local discriminator structure with self-regularization to preserve content features and improve perceptual quality consistently. By constructing a large low-light image quality assessment dataset, Chen \textit{et al.} proposed to enhance the low-light image towards a better visual quality~\cite{chen2023gap}. Although these methods could achieve promising performance for single image enhancement, there is still much room to explore when considering the inter-view correlation for multi-view low-light image enhancement.

Besides, similar efforts are dedicated to multi-view/multi-frame based low-light enhancement. In contrast to conventional cameras,  light-field cameras enable the acquisition of images in a multi-view manner \cite{lamba2020harnessing}. To enhance the light-field image captured in low-light conditions, Lamba \textit{et al.} \cite{lamba2020harnessing} first proposed a two-stage deep neural network, where the global representation block followed by view reconstruction block was designed for low-light light-field view restoration. Wang \textit{et al.} \cite{wang2023multi} proposed a multi-stream progressive restoration network, by which, visual information in different views can be fused and synthesized for the final enhancement. Different from \cite{lamba2020harnessing, wang2023multi} exploiting multi-view aggregation simply via feature concatenation, we perform cross-view feature alignment with adaptive fusion for multi-view feature extraction and aggregation in different views. For the low-light stereo images, a dual-view enhancement network based on the Retinex theory was proposed in \cite{huang2022low}, which was characterized by a coarse-to-fine restoration. Compared to \cite{huang2022low}, we design a recurrent collaborative network to iteratively perform intra-view enhancement and inter-view alignment and fusion for multi-view image enhancement, by which, the image can be refined in each recurrence in a more careful way. 

In addition to low-light light-field and stereo images, different view information can also be obtained from video frames, known as the low-light video enhancement \cite{triantafyllidou2020low, chhirolya2022low, zheng2022semantic, azizi2022salve, xu2023deep}. Chhirolya \textit{et al.} \cite{chhirolya2022low} designed a self-cross dilated attention module to exploit the inter-frame information. Zheng \textit{et al.} \cite{zheng2022semantic} devised a semantic-guided zero-shot low-light enhancement network, facilitating the video restoration without relying on rigorously paired data. Compared to the above methods, we adopt the multi-view low-light triplets as input and perform feature extraction, enhancement, alignment, and fusion between intra-view or inter-view images, which is different from the domain mapping \cite{triantafyllidou2020low} or zero-shot learning \cite{zheng2022semantic}. 
Moreover, unlike the approach of extending the keyframe enhancement mapping to the remaining frames \cite{azizi2022salve} or only using a single iteration \cite{chhirolya2022low}, we perform individual view enhancements that benefit from multi-view collaboration in a recurrent way. In this work, we systematically study the multi-view low-light image enhancement by constructing a dedicated dataset and designing an effective algorithm that utilizes the cross-view feature correspondence in multi-view collaboration.

\section{The MVLT Dataset} \label{sec3-dataset}
% 说明MVLT数据集与现有低光数据集的区别+构建MVLT的两点根据
In contrast with the popular low-light datasets \cite{wei2018deep, yang2021sparse, lv2021attention} focusing mainly on scene diversity with only one viewpoint available in one scene, the proposed MVLT dataset is specifically established to explore the combination of low-light scene and multi-view representation (\textit{i.e.}, view diversity). Herein, we introduce how the dataset is constructed from the perspectives of multi-view selection and low-light synthesis. 
% popular low-light datasets {chen2018learning}

\begin{figure*}[t]
\centering
\includegraphics[width=\linewidth]{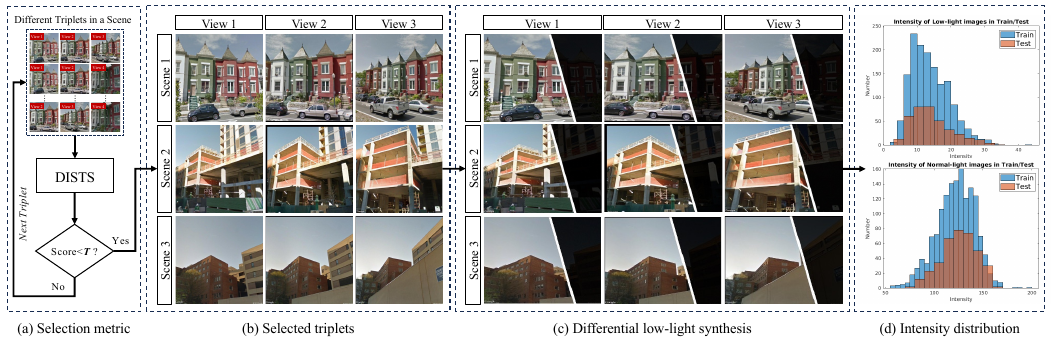}
\caption{Illustration of our MVLT dataset construction and statistics: (a) we adopt the DISTS metric to compute the similarity score with the threshold \textbf{\textit{T}} for multi-view triplets selection; (b) the example triplets of normal-light images; (c) the differential low-light synthesis is composed of brightness reduction and noise simulation; (d) the intensity distribution of low/normal-light images in training and testing set, respectively. Please zoom in for a better visualization.}
\label{fig-dataset-statistics}
\end{figure*}

% 多视角数据的选择
\textit{Selection of multi-view triplets.} We collect multi-view images from the popular object-centric street view dataset \cite{zamir2016generic}, including a large amount of capturing poses and city scenes. In this dataset, every 2$\sim$7 corresponding street view images share the same physical target point, which also indicates the same scene could be captured from 2$\sim$7 different viewpoints. However, there are repeated scenes with large content overlap or low image similarity even among the view groups. To tackle these problems, we employ the Deep Image Structure and Texture Similarity (DISTS) metric \cite{ding2020image} to evaluate image similarity in 2$\sim$7 viewpoints, as depicted in Fig.\ref{fig-dataset-statistics} (a). A lower DISTS score means higher similarity between two images. We empirically set a similarity threshold \textbf{\textit{T}} of 0.2 to select multi-view triplets, ensuring that the similarity score of any two randomly selected images is below 0.2. Furthermore, we filter out the repeated scenes with little viewpoint changes/large content overlap manually. 
Finally, we can obtain 1,860 normal-light street images, \textit{i.e.}, 620 triples of multi-view images. The sampled triples are shown in Fig.~\ref{fig-dataset-statistics} (b). These images are further randomly divided into 1,488 images/496 triples for the training set and 372 images/124 triples for the testing set. All the multi-view images are with a resolution of ${640\times640\times3}$.

% 给选择的多视角数据合成low-light流程：dark+noise两步
\textit{Differential low-light synthesis.} These selected multi-view street triplets serve as normal-light ground truth. In analogous to the procedure of low-light synthesis in \cite{lv2021attention, huang2022low}, we adopt brightness reduction followed by noise simulation to synthesize corresponding low-light images. More specifically, we use linear scaling and gamma transformation to darken multi-view normal-light images via
\begin{equation}
\label{eq1}
\hat{x}_{n}=\beta\times(\alpha\times\hat{\mathcal{R}}_{n})^{\gamma},
\end{equation} 
where $\hat{x}_{n}$ and $\hat{\mathcal{R}}_{n}$ are the synthesized low/normal-light images, respectively, $\alpha$ and $\beta$ denote the linear scaling factors sampled from uniform distributions $U(0.9, 1)$ and $U(0.1, 0.3)$, respectively. And $\gamma$ means the gamma correction sampled from $U(1.4, 2.5)$. Subsequently, the Gaussian-Poisson mixed noise model is integrated into the in-camera processing (ISP) \cite{guo2019toward}, to simulate as realistic noise distribution as possible. It is worth mentioning that due to the viewpoint changes, each single view in the triplet tends to be captured differently. Therefore, the random strategy of parameter sampling during the synthesis pipeline is adopted in a multi-view triplet. Examples are also shown in Fig.~\ref{fig-dataset-statistics} (c).
%The detailed parameters of darkness and noise synthesis are listed in the supplementary material.

\textit{Dataset statistics.} As shown in Fig.\ref{fig-dataset-statistics} (d), we report the intensity distribution of low/normal-light images in training and testing set, respectively. We can derive that the low-light images (or the normal-light counterparts) in training and testing sets share similar distributions. Moreover, both of low-light and normal-light samples cover a large intensity ranges as close to real-word scenes as possible.

\section{The Proposed Approach} \label{sec4-method}
\subsection{Problem Formulation}
Multi-view low-light enhancement aims at restoring normal-light images in collaboration with several other views in the same low-light scene. Herein, we adopt three different views in the multi-view scene. Generally speaking, three dark images from different viewpoints are represented as the set $\mathcal{D}=\{\textbf{\textit{x}}|\textbf{\textit{x}}=(x_{1}, x_{2}, x_{3})\}$ with a common scenario \textbf{\textit{x}}. And these three images have the same spatial width $W$ and height $H$, \textit{i.e.}, $x_{n}\in \mathbb{R}^{W\times H\times 3}, n=1,2,3$. Formally, let $G_e(\cdot)$ denotes the enhancement mapping function, then the restored image $\mathcal{R}_{n}\in \mathbb{R}^{W\times H\times 3}$ can be obtained by,
\begin{equation}
\label{eq1}
\mathcal{R}_{n}=G_{e}(\mathcal{D};\theta), \forall n\in\{1,2,3\},
\end{equation}
where $\theta$ means the learnable network parameters of $G_{e}(\cdot)$, and $n$ represents a random view in the set $\mathcal{D}$. Among the view set, we denote the low-light view to be enhanced as \textit{primary view}, and the other two views are named by \textit{auxiliary views}. For example, when the dark view $x_{2}$ is assumed to be the \textit{primary view}, the \textit{auxiliary views} would contain $x_{1}$ and $x_{3}$ with the enhancement result $\mathcal{R}_{2}$.

The core of $G_{e}(\cdot)$ is to learn the \textit{primary view} enhancement mapping in cooperation with \textit{auxiliary views}. Thus, in order to achieve the desired result as close to the normal-light version as possible, the optimization process could be depicted as
\begin{equation}
\label{eq2}
\hat{\theta}=\mathop{\arg}\mathop{\min}_{\theta} L_{{\rm total}}(\mathcal{R}_{n}, \hat{\mathcal{R}}_{n}),
\end{equation}
where $\hat{\theta}$ is the final optimal network parameters of $G_{e}$ trained by minimizing the total loss $L_{{\rm total}}$, and $\hat{\mathcal{R}}_{n}\in \mathbb{R}^{W\times H\times 3}$ is the normal-light ground truth of \textit{primary view}.

\begin{figure*}[t]
\centering
\includegraphics[width=\linewidth]{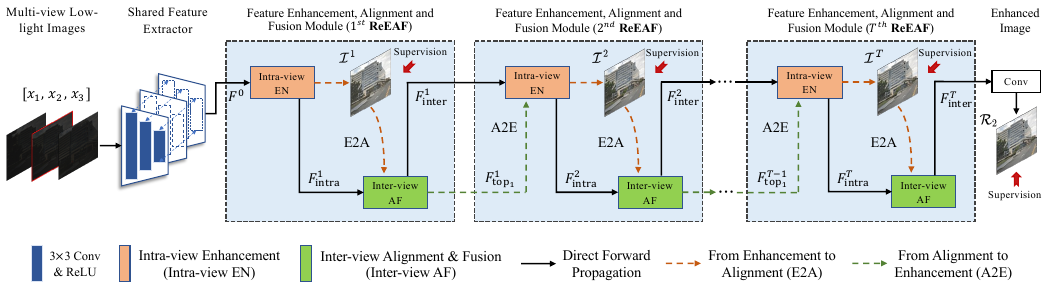}
\caption{Illustration of our proposed multi-view low-light enhancement framework: (i) the multi-view low-light images are grouped into a triplet $\mathcal{D}$ including a \textit{primary view} ($x_{2}$) and two \textit{auxiliary views} ($x_{1}$ and $x_{3}$); (ii) a shared encoder is utilized as the multi-scale feature extractor to obtain multi-view features in different scales from three low-light input views; (iii) the recurrent feature enhancement, alignment and fusion (ReEAF) module is embedded to integrate \textit{primary view} features via multi-view collaboration. In each recurrent unit, the ReEAF is composed of Intra-view Enhancement (Intra-view EN) followed by Inter-view Alignment and Fusion (Inter-view AF); and (iv) the finally enhanced image $\mathcal{R}_{2}$ corresponding to \textit{primary view} $x_{2}$ could be produced by a single convolutional layer at the end of the fusion stage.}
\label{fig3-framework}
\end{figure*}

\subsection{Overview of the Proposed Method}
%%% 与Fig3标题分四步说明网络的主要组成稍微不同，这里用一两句话简要说明前两种组成, 即多视角输入+shared特征提取encoder, 后面重点说明ReEAF的设计inspiration和主要构成，结尾一句话说明产生enhanced image
As shown in Fig.~\ref{fig3-framework}, our proposed enhancement framework takes multi-view low-light triplet as input and produces the enhanced \textit{primary view} in an end-to-end manner. More specifically, given a triplet of multi-view low-light images including a \textit{primary view} $x_{2}$ and two \textit{auxiliary views} $x_{1}$ and $x_{3}$, we first adopt a shared multi-scale feature extractor to obtain the multi-view features in different scales, from which diverse contextual information across scales could be effectively captured. 
%Motivated by the property of similar example recurrence in a single image \cite{zhou2020cross, li2020mucan}, the \textit{primary view} restoration benefits not only from attentive feature sampling within a single view but also the multi-view features for aggregating similar patches from the \textit{auxiliary views} in the same scene. 
Herein, the \textit{primary view} feature is expected to be enhanced from \textit{primary view} itself and two corresponding \textit{auxiliary views} interactively. Along this vein, the Recurrent feature Enhancement-Alignment-Fusion (ReEAF) module is designed to facilitate \textit{primary view} via multi-view collaboration.

In each recurrent unit, the ReEAF is composed of Intra-view Enhancement (Intra-view EN) followed by Inter-view Alignment and Fusion (Inter-view AF). In the Intra-view EN, we impose spatial and channel feature enhancement on each single view. Regarding the Inter-view AF, we first perform the feature alignment between the two \textit{auxiliary views} and the \textit{primary view}, then the feature fusion is conducted across different views. We connect the Intra-view EN and Inter-view AF by two interaction strategies, \textit{i.e.}, from enhancement to alignment (E2A) and from alignment to enhancement (A2E). The design details are elaborated as follows.

%subsequently the multi-view features are integrated to generate efficient \textit{primary view} feature in the Inter-view AF. To make use of the multi-view collaboration, two different interactions from enhancement to alignment (E2A) and  from alignment to enhancement (A2E) are introduced between intra-view and inter-view feature propagation. Specifically, the E2A integrates the supervised \textit{primary view} image $\mathcal{I}^{t}$ into Inter-view AF, which can guide the quality-aware feature weighting for the \textit{top-k} most similar patches $\mathcal{F}_{{\rm top-k}}^{t}$ across different views. Herein, the $t$ indicates $t^{th}$ Inter-view AF in our RCNet. In contrast, the A2E propagates the sorted \textit{top-1} most similar feature $\mathcal{F}_{{\rm top-1}}^{t-1}$ to Intra-view EN in the next recurrent unit for attentive spatial sampling. Finally, a single convolutional layer with 3$\times$3 kernel is utilized to generate the enhanced image $\mathcal{R}_{2}$ at the end of fusion stage.

%In the followings, we will elaborate how to develop the ReEAF module, and therein present how we build the collaboration mechanism from enhancement (alignment) to alignment (enhancement), \textit{i.e.}, E2A(A2E). Without loss of generality, we take the $t^{th}$ ($t=2,...,T-1$) ReEAF module as an example for feature propagation, where $T$ indicates the maximum number of the recurrent unit ReEAF, as shown in Fig.~\ref{fig4-intra-inter}.

\subsection{Intra-view Enhancement (Intra-view EN)} \label{Intra-view EN}
Supposing the output of the multi-scale feature extractor is $\boldsymbol{F}^{0} \in \mathbb{R}^{W\times H\times C\times 3}$ and the function of the Intra-view EN is $G_{{\rm intra}}^{t}(\cdot)$, then the enhanced feature $\boldsymbol{F}_{{\rm intra}}^{t} \in \mathbb{R}^{W\times H\times C\times 3}$ at the $t$-th ReEAF can be obtained by,
\begin{equation}\label{eq3-intra}
\boldsymbol{F}_{{\rm intra}}^{t} = \left\{
\begin{array}{ll}
G_{{\rm intra}}^{1}(\boldsymbol{F}^{0}) & (t=1),\\
G_{{\rm intra}}^{t}(\boldsymbol{F}_{{\rm inter}}^{t-1}, \boldsymbol{F}_{{\rm {top}_{1}}}^{t-1}) & (t>1),
\end{array} \right.
\end{equation}
where the $\boldsymbol{F}_{{\rm inter}}^{t-1} \in \mathbb{R}^{W\times H\times C\times 3}$ and $ \boldsymbol{F}_{{\rm {top}_{1}}}^{t-1}$ are the output features of the Inter-view AF and A2E modules, which we would elaborate in subsection \ref{Inter-view AF} and \ref{A2E}, respectively. As shown in Fig.~\ref{fig4-intra-inter}, the $G_{{\rm intra}}^{t}(\cdot)$ consists of both a spatial attention branch $G_{{\rm spatial}}^{t}(\cdot)$ and a channel attention branch $G_{{\rm channel}}^{t}(\cdot)$. In particular, the spatial attention aims to capture the enhancement levels in different regions as the illumination degradation is not uniformly distributed. In the first stage ($t$ = 1), the attention is generated from each single view itself. In the following stages ($t >$ 1), we further introduce  $\boldsymbol{F}_{{\rm {top}_{1}}}^{t-1}$ for the attention generation. The  $\boldsymbol{F}_{{\rm {top}_{1}}}^{t-1}$ is formed by the feature patches searched in each single view that share the most (top 1) similarity with the \textit{primary view}. 
%The assumption behind the utilization of cross-view information lies in that 
Herein, the utilization of cross-view information highly benefits spatial attention estimation as it provides a measurement of the effort that we should pay for the enhancement of each region. For example, more attention should be paid to the regions whose most similar regions are still with unpleasant quality.
%the enhancement gains that the primary view can be obtained from other views and more attention should be distributed to the regions that benefit less.
For the channel attention branch, a squeeze-and-excitation operation is adopted to collect the contextual information in the whole intra-view feature maps. Finally, we  treat the attention-based enhanced features as a residue of the initial one and obtain the final enhanced features as the input of the E2A and Inter-view AF modules, which can be formulated as follows,
\begin{equation}\label{}
\begin{aligned}
G_{{\rm intra}}^{t}(\boldsymbol{F}_{{\rm inter}}^{t-1}, \boldsymbol{F}_{{\rm {top}_{1}}}^{t-1}) = & G_{{\rm spatial}}^{t}(\boldsymbol{F}_{{\rm inter}}^{t-1}, \boldsymbol{F}_{{\rm {top}_{1}}}^{t-1}) \\
& \otimes G_{{\rm channel}}^{t}(\boldsymbol{F}_{{\rm inter}}^{t-1}) \oplus \boldsymbol{F}_{{\rm inter}}^{t-1},
\end{aligned}
\end{equation}
%and
%\begin{equation}\label{}
%G_{{\rm spatial}}^{t}(\boldsymbol{F}_{{\rm inter}}^{t-1}, \boldsymbol{F}_{{\rm {top}_{1}}}^{t-1}) = %f_{{\rm aconv}}^{t}(\boldsymbol{F}_{{\rm inter}}^{t-1}) \otimes \boldsymbol{F}_{{\rm {top}_{1}}}^{t-1},
%\end{equation}
%where the $f_{{\rm aconv}}^{t}$ denotes several asymmetric convolutions for hierarchical feature extraction, and 
where the operators $\otimes$ and $\oplus$ mean the element-wise multiplication and addition, respectively. 

\subsection{Enhancement to Alignment (E2A)} \label{E2A}
\begin{figure*}[t]
\centering
\includegraphics[width=\linewidth]{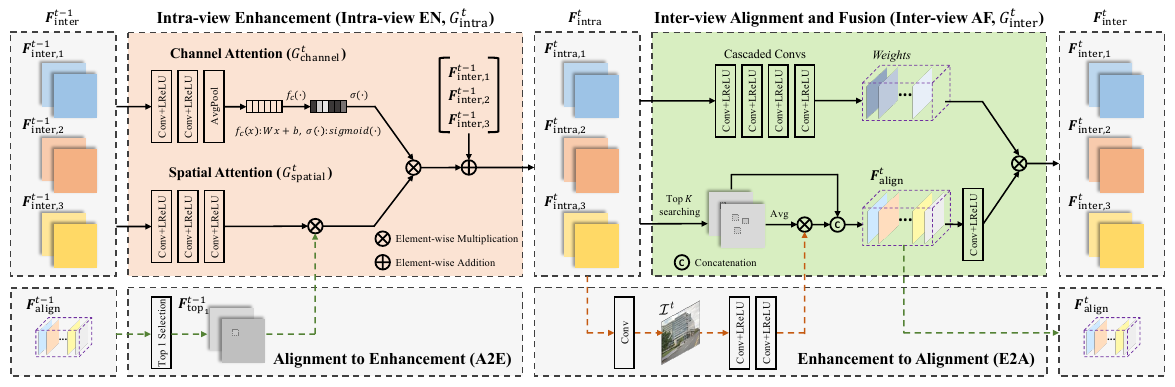}
\caption{Illustration of the recurrent feature enhancement-alignment-fusion (ReEAF) module.}
\label{fig4-intra-inter}
\end{figure*}

Given the $\boldsymbol{F}_{{\rm intra}}^{t}$, the E2A aims to predict the enhanced images $\mathcal{I}^{t} \in \mathbb{R}^{W\times H\times 3}$ at the stage $t$ for the Inter-view AF. As shown in Fig.~\ref{fig4-intra-inter}, the image predictor only consists of one convolutional layer with the kernel 3$\times$3. Herein, the image predictor plays two roles in our method: 1) Supervised by the normal-light image $\hat{\mathcal{R}}_{2}$ (\textit{primary view}), the multi-stage guidance leads to a more precious enhancement. 2) The prediction result $\mathcal{I}^{t}$ bridges the Intra-view EN and Inter-view AF by providing a confidence map in the Inter-view AF. In the Inter-view AF, the features in different views are first aligned with the \textit{primary view} by searching top K similar patches, then the fusion is conducted to aggregate the aligned features. However, the returned patches may not be reliable especially when the quality of the regions of the \textit{primary view} is degraded severely, as such, the top K patches should be fused with different confidences. Herein, we adopt the  $\mathcal{I}^{t}$ to estimate the fusion confidence auxiliarly, as the quality degradation can be well reflected by the prediction result. More details regarding the utilization of $\mathcal{I}^{t}$ in  Inter-view AF would be described in subsection \ref{Inter-view AF}.

%For simplification, we show one intermediate recurrent unit, including (a) Intra-view Enhancement (Intra-view EN) and (b) Inter-view Alignment and Fusion (Inter-view AF), wherein lies the interactions from enhancement (alignment) to alignment (enhancement), \textit{i.e.}, E2A(A2E). The figure is best viewed in color.

\subsection{Inter-view Alignment and Fusion (Inter-view AF)} \label{Inter-view AF}
\begin{figure}[t]
\centering
\includegraphics[width=0.96\linewidth]{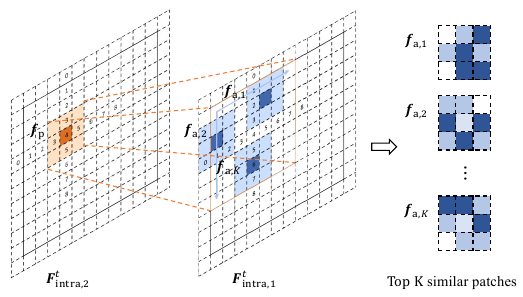}
\caption{Illustration of the feature alignment by searching top K similar patches in a local region.}
\label{fig4-topk}
\end{figure}

Based upon the $\boldsymbol{F}_{{\rm intra}}^{t}$ and $\mathcal{I}^{t}$, the Inter-view AF $G_{{\rm inter}}^{t}(\cdot)$ aims to explore the favorable features $\boldsymbol{F}_{{\rm inter}}^{t}$ in cross-views for the \textit{primary view} enhancement,
\begin{equation}\label{}
\left\{
\begin{array}{rll}
\boldsymbol{F}_{{\rm {top}_{1}}}^{t}, \boldsymbol{F}_{{\rm inter}}^{t} =& G_{{\rm inter}}^{t}(\boldsymbol{F}_{{\rm intra}}^{t}, \mathcal{I}^{t}) & (t<T),\\
\boldsymbol{F}_{{\rm inter}}^{T} =& G_{{\rm inter}}^{T}(\boldsymbol{F}_{{\rm intra}}^{T}, \mathcal{I}^{t}) & (t=T),
\end{array} \right.
\end{equation}
To achieve this, two steps are included in our Inter-view AF, \ie, cross-view feature alignment and adaptive fusion. In the cross-view feature alignment, the texture recurrences in cross-views are mined in a patch-level for the \textit{primary view} \cite{yan2020deep}. As can be seen in Fig.~\ref{fig4-intra-inter}, given the primary feature $\boldsymbol{F}_{{\rm intra, 2}}^{t} \in \mathbb{R}^{W\times H\times C}$ and the two auxiliary features $\boldsymbol{F}_{{\rm intra, 1}}^{t} \in \mathbb{R}^{W\times H\times C}$ and $\boldsymbol{F}_{{\rm intra, 3}}^{t} \in \mathbb{R}^{W\times H\times C}$, we first partition those features into non-overlap patches with the patch size set to 7$\times$7. Taking aligning the  $\boldsymbol{F}_{{\rm intra, 1}}^{t}$ to $\boldsymbol{F}_{{\rm intra, 2}}^{t}$ as an example shown in Fig.~\ref{fig4-topk}, supposing one patch feature in $\boldsymbol{F}_{{\rm intra, 2}}^{t}$ is denoted as  $\boldsymbol{f}_{{\rm p}}$, we find its top K nearest neighbors (denoted as $\boldsymbol{f}_{{\rm a},1}$,  $\boldsymbol{f}_{{\rm a},2}$,$\ldots$, $\boldsymbol{f}_{{\rm a},K}$) on $\boldsymbol{F}_{{\rm intra, 1}}^{t}$ within a local search area and their correlation $\rho$ is computed as the normalized inner product,
\begin{equation}\label{eq:structure}
\rho(\boldsymbol{f}_{{\rm p}},\boldsymbol{f}_{{\rm a, i}})= \frac{\boldsymbol{f}_{{\rm p}}^T{\boldsymbol{f}_{{\rm a, i}}}}{\left \|\boldsymbol{f}_{{\rm p}}\right \|\left \|\boldsymbol{f}_{{\rm a, i}} \right \|}  \quad 
i = 1,2,\ldots, K.
\end{equation}
Based upon the searched top K most correlated patches, we herein do not fuse those patches directly, as their similarity may not be reliable due to quality degradation. To account for this, we further calculate their average result $\boldsymbol{f}_{{\rm avg}}$ as a complementary candidate and weight it by the confidence map estimated by the $\mathcal{I}^{t}$ as follows,
\begin{equation}\label{eq:avg}
\begin{aligned}
\overline{\boldsymbol{f}}_{{\rm avg}} &= \boldsymbol{C}(p)*\boldsymbol{f}_{{\rm avg}}\\
   &= \frac{\boldsymbol{C}(l)}{K} (\boldsymbol{f}_{{\rm a, 1}} \oplus \boldsymbol{f}_{{\rm a, 2}},\ldots, \oplus \boldsymbol{f}_{{\rm a, K}}),
\end{aligned}
\end{equation}
and 
\begin{equation}\label{eq:conf}
\boldsymbol{C} = G_{{\rm cof}}^{t}(\mathcal{I}^{t}),
\end{equation}
where $\boldsymbol{C}$ is the confidence map, $ G_{{\rm cof}}^{t}(\cdot)$ is the confidence evaluator consisting of convolutional layers, and $l$ is the spatial index (location) of the $\boldsymbol{f}_{{\rm p}}$ in $\boldsymbol{F}_{{\rm intra, 2}}^{t}$. Subsequently, we concatenate all those candidates along the channel dimension to obtain the final aligned feature, 
\begin{equation}\label{eq:conc}
\boldsymbol{F}_{{\rm align, 1}}^{t} = \left[\boldsymbol{f}_{{\rm a, 1}}, \boldsymbol{f}_{{\rm a, 2}}, \ldots ,\overline{\boldsymbol{f}}_{{\rm avg}}\right],
\end{equation}
where $\boldsymbol{F}_{{\rm align, 1}}^{t} \in \mathbb{R}^{W\times H\times C}$ is the aligned results between the $\boldsymbol{F}_{{\rm intra, 1}}^{t}$ and $\boldsymbol{F}_{{\rm intra, 2}}^{t}$ and the $\left[ \cdot \right]$ represents the concatenation operation. Thus the adaptive fusion can be as follows,
\begin{equation}\label{}
\begin{aligned}
\boldsymbol{F}_{{\rm inter}}^{t} &= G_{{\rm wt}}^{t}(\left[\boldsymbol{F}_{{\rm intra, 1}}^{t}, \boldsymbol{F}_{{\rm intra, 2}}^{t},\boldsymbol{F}_{{\rm intra, 3}}^{t} \right]) \\ 
&\otimes G_{{\rm conv}}(\left[\boldsymbol{F}_{{\rm align, 1}}^{t},\boldsymbol{F}_{{\rm align, 2}}^{t},\boldsymbol{F}_{{\rm align, 3}}^{t}\right]),
\end{aligned}
\end{equation}
where $G_{{\rm wt}}^{t}(\cdot)$ indicates the weight prediction function, consisting of four convolutional layers. Analogous to the $\boldsymbol{F}_{{\rm align, 1}}^{t}$, the $\boldsymbol{F}_{{\rm align, 3}}^{t}$ is the aligned result between the $\boldsymbol{F}_{{\rm intra, 3}}^{t}$ and $\boldsymbol{F}_{{\rm intra, 2}}^{t}$, and the $\boldsymbol{F}_{{\rm align, 2}}^{t}$ is the aligned result from the $\boldsymbol{F}_{{\rm intra, 2}}^{t}$ itself. The $G_{{\rm conv}}$ is a process function for their concatenation result.

\subsection{Alignment to Enhancement (A2E)} \label{A2E}
In the Inter-view AF, we obtain the top K most similar candidates from each view for $\boldsymbol{F}_{{\rm intra, 2}}^{t}$. According to our description in subsection \ref{Intra-view EN}, the $\boldsymbol{F}_{{\rm {top}_{1}}}^{t}$ is utilized for more accurate attention estimation in Intra-view EN. Herein, the $\boldsymbol{F}_{{\rm {top}_{1}}}^{t}$ is formed by the searched feature patches that share the most (top 1) similarity with the \textit{primary view}. For example, $\boldsymbol{F}_{{\rm {top}_{1}},1}^{t} (l) = \boldsymbol{f}_{{\rm a, 1}}$ ($l$ is an arbitrary spatial index  in $\boldsymbol{F}_{{\rm intra, 2}}^{t}$) when those top K patches are from $\boldsymbol{F}_{{\rm intra, 1}}^{t}$. Analogously, we could obtain  $\boldsymbol{F}_{{\rm {top}_{1}},2}^{t}$ and  $\boldsymbol{F}_{{\rm {top}_{1}},3}^{t}$ and
\begin{equation}\label{eq:cont}
\boldsymbol{F}_{{\rm  {top}_{1}}}^{t} =\left[ \boldsymbol{F}_{{\rm {top}_{1}},1}^{t}, \boldsymbol{F}_{{\rm {top}_{1}},2}^{t} , \boldsymbol{F}_{{\rm {top}_{1}},3}^{t} \right].
\end{equation}
Finally, we deliver the $\boldsymbol{F}_{{\rm {top}_{1}}}^{t}$ to the $(t+1)$-th Intra-view EN, thus the Intra-view EN and Inter-view AF are connected in series. In summary, the Intra-view EN, E2A, Inter-view AF, and A2E are subsequently linked and form a full ReEAF unit module. In our method, three ReEAF units are cascaded and enhance the image at the \textit{primary view} in an iterative way.

{\linespread{1.4}
\begin{table*}[t]
    \centering
    {\linespread{1.0}
    \caption{Comparison of quantitative results in terms of PSNR, SSIM, FSIM, VIF, and LOE on the MVLT dataset. The arrow $\uparrow$($\downarrow$) behind quality metrics means that the larger(smaller) value is better. The values highlighted with bold font and underline indicate ranking the first and second place, respectively.}
    \label{tab-result}} % \label{tab-result}要放在{\linespread{1.0}}中去，否则会出现Table序号出现问题
    \scalebox{0.84}{ % 0.82
    \begin{threeparttable} %添加此处
        \begin{tabular}{c c}
            % First table
            \begin{tabular}{l|l|l|l|l|l|l}
            \toprule
             Category & Method & PSNR$\uparrow$ & SSIM$\uparrow$ & FSIM$\uparrow$ & VIF$\uparrow$ & LOE$\downarrow$ \\
            \midrule
             \multirow{7}{*}{\shortstack{Single-based \\ Methods \\ (Traditional)}}
             & Dong\cite{dong2010fast} & 14.59 & 0.4876 & 0.8397 & 0.3119 & \underline{290.0} \\ %2010'
             & NPE\cite{wang2013naturalness} & \underline{17.45} & 0.5001 & \underline{0.8495} & 0.3609 & 328.5  \\ %2013'
             & LIME\cite{guo2016lime} & 17.28 & 0.4709 & 0.8183 & \underline{0.3754} & 653.2         \\ %2016'
             & SRIE\cite{fu2016weighted} & 9.75 & 0.4264 & 0.8161 & 0.3362 & 309.3       \\ %2016'CVPR
             & BIMEF\cite{ying2017bio} & 11.37 & 0.5308 & 0.8218 & 0.3388 & 328.8        \\ %2017'
             & JieP\cite{cai2017joint} & 9.94 & 0.4438 & 0.8259 & 0.3389 & 343.2         \\ %2017'ICCV
             & RRM\cite{li2018structure} & 11.06 & 0.5858 & 0.7531 & 0.2763 & 300.1      \\ %2018'TIP
            \midrule
             & RCNet (Ours) & \textbf{26.45} & \textbf{0.8844} & \textbf{0.9397} & \textbf{0.4594} & \textbf{124.8} \\ %2023'
            \bottomrule
             Category & Method & PSNR$\uparrow$ & SSIM$\uparrow$ & FSIM$\uparrow$ & VIF$\uparrow$ & LOE$\downarrow$ \\
            \midrule
             \multirow{7}{*}{\shortstack{Multi-based \\ Methods \\ (Deep)}}
             & SALVE\cite{azizi2022salve} & 10.86  & 0.5400  & 0.7392  & 0.2111  & 370.0        \\ %2023'APSIPA
             & Chhirolya\cite{chhirolya2022low} & 15.70  & 0.5387  & 0.5523  & 0.0093 & 815.6    \\ %2022'BMVC
             & SGZSL\cite{zheng2022semantic} & 16.58  & 0.5323  & 0.8423  & 0.3185  & 547.7       \\ %2022'WACV
             & DP3DF\cite{xu2023deep} & 22.83  & 0.7448  & 0.8618  & 0.1811  & 261.1       \\ %2023'AAAI
             % & SIDGAN\cite{triantafyllidou2020low} & - & - & - & - & -      \\ %2020'ECCV
             & L3Fnet\cite{lamba2020harnessing} & 21.48  & 0.8149  & 0.9007  & 0.3759  & 307.4    \\ %2020'TIP
             & MSPnet\cite{wang2023multi} & 19.90  & 0.8170  & 0.8963  & 0.3854  & 349.1    \\ %2023'TCI
             & DVENet\cite{huang2022low} & \underline{26.03}  & \underline{0.8468}  & \underline{0.9265} & \underline{0.4074}  & \underline{156.1}       \\ %2022'TMM
            \midrule
             & RCNet (Ours) & \textbf{26.45} & \textbf{0.8844} & \textbf{0.9397} & \textbf{0.4594} & \textbf{124.8} \\ %2023'
            \bottomrule
            \end{tabular}
        &
            % Second table
            \begin{tabular}{l|l|l|l|l|l|l}
            \toprule
             Category & Method & PSNR$\uparrow$ & SSIM$\uparrow$ & FSIM$\uparrow$ & VIF$\uparrow$ & LOE$\downarrow$ \\
            \midrule
             \multirow{17}{*}{\shortstack{Single-based \\ Methods \\ (Deep)}}
             & RetinexNet\cite{wei2018deep} & 15.88 & 0.4384 & 0.7905 & 0.2486 & 709.0  \\ %2018'BMCV
             & MBLLEN\cite{lv2018mbllen} & 17.20 & 0.7119 & 0.9084 & 0.4293 & 230.2      \\ %2018'2018
             & KinD\cite{zhang2019kindling} & 23.29 & 0.8731 & 0.9339 & 0.4337 & 238.9   \\ %2019'ACMMM
             & DLN\cite{wang2020lightening} & 22.19 & 0.7502 & 0.8929 & 0.3777 & 213.5   \\ %2020'TIP
             & ZeroDCE\cite{guo2020zero} & 15.71 & 0.5150 & 0.8233 & 0.3052 & 757.0     \\ %2020'CVPR
             & LPNet\cite{li2020luminance} & 18.85 & 0.8060 & 0.8768 & 0.3899 & 156.0    \\ %2020'TMM
             & DSLR\cite{lim2020dslr} & 23.34 & 0.7927 & 0.8811 & 0.3182 & 246.6         \\ %2020'TMM
             & EnGAN\cite{jiang2021enlightengan} & 19.82 & 0.6918 & 0.8790 & 0.3845 & 415.5 \\ % 2021'TIP
             & RUAS\cite{liu2021retinex} & 14.90 & 0.4827 & 0.7915 & 0.2983 & 689.5      \\ %2021'CVPR
             % & ZeroDCE++\cite{li2021learning} & 16.45 & 0.5315 & 0.8429 & 0.3170 & 542.7\\ %2021'TPAMI
             & DRBN\cite{yang2020fidelity} & 23.05 & 0.8106 & 0.9007 & 0.2568 & 287.9    \\ %2020'CVPR
             & MIRNet\cite{zamir2020learning} & 25.05 & 0.8560 & 0.9247 & 0.4264 & 164.9 \\ %2020'ECCV
             & Uformer\cite{wang2022uformer} & 23.14 & 0.8051 & 0.9128 & 0.4017 & 257.2  \\ %2021'CVPR
             & SGM\cite{yang2021sparse} & 23.73 & 0.8692 & 0.9283 & 0.4342 & 239.1       \\ %2021'TIP
             & LLFlow\cite{wang2022low} & 25.54 & 0.8511 & 0.9242 & 0.3706 & 215.9       \\ %2022'AAAI
             & SNR\cite{xu2022snr} & 25.72 & 0.8733 & \underline{0.9388} & \underline{0.4506} & \textbf{124.3}   \\ %2022'CVPR
             & MBPNet\cite{zhang2023multi} & 23.70 & 0.6680 & 0.9011 & 0.3928 & 200.9   \\ %2023'TIP
             & LIVENet\cite{makwana2024livenet} & \underline{25.88} & \underline{0.8737} & 0.9337 & 0.4020 & 159.7   \\ %2024'WACV
            \midrule
             & RCNet (Ours) & \textbf{26.45} & \textbf{0.8844} & \textbf{0.9397} & \textbf{0.4594} & \underline{124.8} \\ %2023'
            \bottomrule
            \end{tabular}  
        \end{tabular}
    \begin{tablenotes}[para,flushleft] %添加此处，“[]”是左对齐的
    \small %添加此处
	\item \textit{Note that Single-based Methods mean the single image based methods, and Multi-based Methods indicate multi-view/multi-frame based methods.} %添加此处
     \end{tablenotes} %添加此处
\end{threeparttable}} %添加此处
\end{table*}}

\subsection{The Loss Function}
%In order to generate high-quality enhanced images as close to normal-light counterparts as possible, 
Our multi-view enhancement network is supervised by the loss function $L_{{\rm total}}$ with inputs of the intermediate result $\mathcal{I}^{t}$ of the $t$-th E2A, the final network output $\mathcal{R}_{n}$ and the corresponding ground truth $\hat{\mathcal{R}}_{n}$ of the \textit{primary view},
\begin{equation}\label{}
L_{{\rm total}} = \sum_{t=1}^{T}L_{{\rm rec}}(\mathcal{I}^{t}, \hat{\mathcal{R}}_{n})+L_{{\rm rec}}(\mathcal{R}_{n}, \hat{\mathcal{R}}_{n}),
\end{equation}
where the reconstruction loss function $L_{{\rm rec}}$ is composed of two components, \textit{i.e.}, the pixel and structure consistency constraints. Specifically, given two images $X$ and $Y$ with the same dimensions, we adopt the $\ell_1$ normalization to calculate the absolute pixel error between the enhanced result $X$ and the ground truth $Y$. We further utilize the Structure SIMilarity Index (SSIM)\cite{wang2004image} to compare the image similarity. To this end, the reconstruction loss $L_{{\rm rec}}$ could be calculated by
\begin{equation}\label{}
L_{{\rm rec}}(X, Y) = \left\|X-Y\right\|_{1}+1-{\rm SSIM}(X, Y).
\end{equation}
%where SSIM measures the similarity of local image patches.
%\begin{equation}\label{}
%{\rm SSIM}(X, Y) = \frac{(2\mu_{X}\mu_{Y}+c_{1})(2\sigma_{XY}+c_{2})}{(\mu_{X}^{2}+\mu_{Y}^{2}+c_{1})(\sigma_{X}^{2}+\sigma_{Y}^{2}+c_{2})},
%\end{equation}
%where $\mu_{X}$ and $\mu_{Y}$, $\sigma_{X}^{2}$ and $\sigma_{Y}^{2}$ are the mean intensities and the variances of $X$ and $Y$, respectively, and $\sigma_{XY}$ is the covariance between $X$ and $Y$. Note that these two small positive constants $c_{1}$ and $c_{2}$ are set for preventing zero division error.

\section{Experimental Results} \label{sec5-experiments}
In this section, we present the experimental results, including experimental settings, performance comparisons, ablation study, application and model complexity discussion.
\begin{figure*}[t]
\centering
\includegraphics[width=\linewidth]{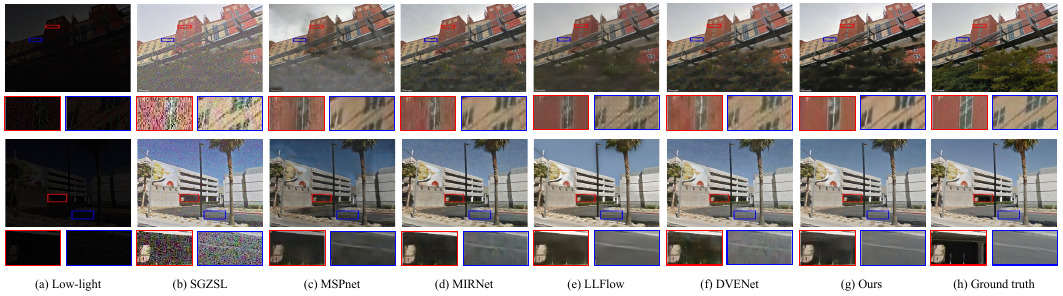}
\caption{Qualitative comparisons of different methods on the MVLT dataset. The selected regions are zoomed in for better visualization.}
\label{fig-syn}
\end{figure*}

\begin{figure*}[t]
\centering
\includegraphics[width=\linewidth]{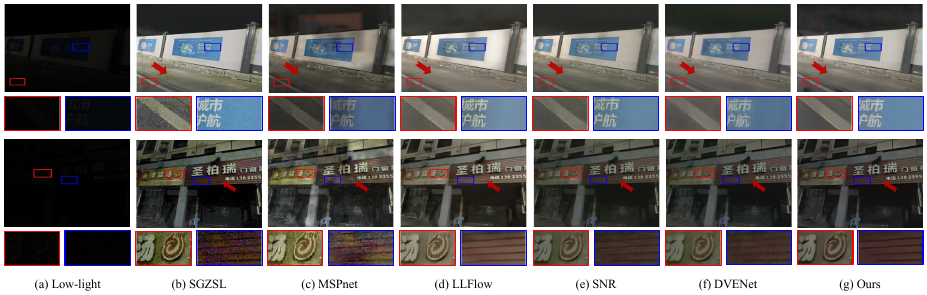}
\caption{Qualitative comparisons of different methods on the real-world scenes. The selected regions are zoomed in for better visualization.}
\label{fig-real}
\end{figure*}

\subsection{Experimental Settings}
\textbf{Implementation Details.} The multi-view enhancement network RCNet is implemented on the Pytorch framework. During network training, random cropping and horizontal flipping are adopted as data augmentation for multi-view images. Therefore, the images in the MVLT dataset are randomly cropped into training patches with the size of 96$\times$96, and then horizontally flipped at a probability of 50\%. The number of training patches in a mini-batch is set to 24, in which one-third is randomly selected as \textit{primary view} and the remaining two-thirds as \textit{auxiliary views}. We use the Adam optimizer for RCNet optimization, and the momentum $\beta_{1}$ and $\beta_{2}$ of Adam are configured with 0.9 and 0.999, respectively. The learning rate is initialized as 2e-4 and decreased to 1e-5 after 37,000 iterations. We further train the whole network to convergence via another 55,000 iterations. During the testing, the multi-view low-light triplet is fed into the network without any cropping, and we could obtain the enhanced \textit{primary view} in an end-to-end manner.
%Pytorch framework \cite{paszke2019pytorch}
%Adam optimizer \cite{kingma2014adam}

\textbf{Benchmarks.} To validate the superiority of our framework, we compare the proposed RCNet with recent state-of-the-art methods, including seven traditional low-light image enhancement algorithms, \textit{i.e.}, Dong \cite{dong2010fast}, NPE\cite{wang2013naturalness}, LIME\cite{guo2016lime}, SRIE\cite{fu2016weighted}, BIMEF\cite{ying2017bio}, JieP\cite{cai2017joint} and RRM\cite{li2018structure}, and seventeen deep single image-based methods, \textit{i.e.}, RetinexNet\cite{wei2018deep}, MBLLEN\cite{lv2018mbllen}, KinD\cite{zhang2019kindling}, DLN\cite{wang2020lightening}, ZeroDCE\cite{guo2020zero, li2021learning}, LPNet\cite{li2020luminance}, DSLR\cite{lim2020dslr}, EnGAN\cite{jiang2021enlightengan}, RUAS\cite{liu2021retinex}, DRBN\cite{yang2020fidelity}, MIRNet\cite{zamir2020learning}, Uformer\cite{wang2022uformer}, SGM\cite{yang2021sparse}, LLFlow\cite{wang2022low}, SNR\cite{xu2022snr}, MBPNet \cite{zhang2023multi} and LIVENet\cite{makwana2024livenet}, as well as seven multi-view/multi-frame based methods, \textit{i.e.}, L3Fnet \cite{lamba2020harnessing}, MSPnet \cite{wang2023multi}, DVENet \cite{huang2022low}, Chhirolya \cite{chhirolya2022low}, SGZSL \cite{zheng2022semantic}, DP3DF \cite{xu2023deep}, and SALVE \cite{azizi2022salve}. For a fair comparison, we use the officially released code of all above methods with their default training and testing settings. In particular, we arrange the multi-view low-light triplet in the same scene as a format of light fields or stereo images for multi-view methods.

\textbf{Evaluation Measures.} In order to evaluate the enhancement performance quantitatively, we use five different quality measures to evaluate the quality of the enhanced result, including Peak Signal-to-Noise Ratio (PSNR), Structure SIMilarity Index (SSIM)\cite{wang2004image}, Feature SIMilarity (FSIM) index \cite{zhang2011fsim}, Visual Information Fidelity (VIF) \cite{sheikh2006image} as well as Lightness-Order-Error (LOE) \cite{wang2013naturalness}. More specifically, PSNR and SSIM put emphasis on pixel-based fidelity and structure-based similarity between the enhanced result and normal-light image, respectively. Since the human visual system (HVS) depends on local features, FSIM calculates the feature similarity by integrating the contrast-invariant phase congruency and the image gradient magnitude complementarily. Furthermore, VIF is developed to measure the visual information fidelity of the resultant image, while LOE is specially designed to quantify the lightness order error for reflecting the naturalness preservation of the enhanced image. In general, larger values of PSNR, SSIM, FSIM and VIF, while smaller value of LOE indicate higher quality of the enhanced image.

\subsection{Performance Comparisons}
\textbf{Quantitative Results.} Table~\ref{tab-result} tabulates the numerical results of the proposed RCNet in comparison with other methods on the MVLT dataset, which are in terms of PSNR, SSIM, FSIM, VIF and LOE. From this table, we can see that our RCNet achieves a favorable performance than recent state-of-the-art methods. More specifically, the traditional methods hardly provide consistent improvements among the five measures due to the limitation of handcraft priors, especially for the LIME\cite{guo2016lime} with the promising noise suppression but poor lightness order. When compared with the deep single-based and multi-based methods, our proposed RCNet obtains significant quality enhancement than the second-best method. In detail, our method performs better than DVENet \cite{huang2022low} by 0.42dB on the PSNR metric, while better than LIVENet\cite{makwana2024livenet} by 0.0107 on the SSIM metric, and furthermore is superior to SNR \cite{xu2022snr} by 0.009 and 0.0088 on the FSIM and VIF metrics, respectively. It can be demonstrated that our RCNet can improve the enhanced results with effective noise suppression and structural details preservation. Although the lightness order value by our method is 0.5 more than SNR\cite{xu2022snr}, we achieve the second-best result in terms of LOE with competitive performance. This is probably because SNR \cite{xu2022snr} uses the additionally estimated signal-noise-ratio map as a prior for guided enhancement, while our method takes only the low-light images as input.
%Furthermore, to compare the quality metrics more intuitively, we adopt the radar maps with five different sub-latitudes to fully demonstrate the superiority of our scheme, as shown in Fig.~\ref{fig7-radar}. In particular, the larger area of the closed pentagon region indicates the better performance of this method. It is apparent from these two radar maps that our method achieves superior performance with the largest areas against the state-of-the-arts.
%\begin{figure}[t]
%\centering
%\includegraphics[width=\linewidth]{Figs/fig7-radar.pdf}
%\caption{Illustration of the radar maps for comparing the proposed RCNet with recent state-of-the-art methods (\textit{Left}: seven traditional algorithms and two Retinex-based methods, \textit{Right}: nine deep learning based methods). It is worth noting that the LOE-axis of radar map is arranged in the opposite direction since smaller value of LOE is better.}
%\label{fig7-radar}
%\end{figure}

\begin{figure*}[t]
\centering
\includegraphics[width=\linewidth]{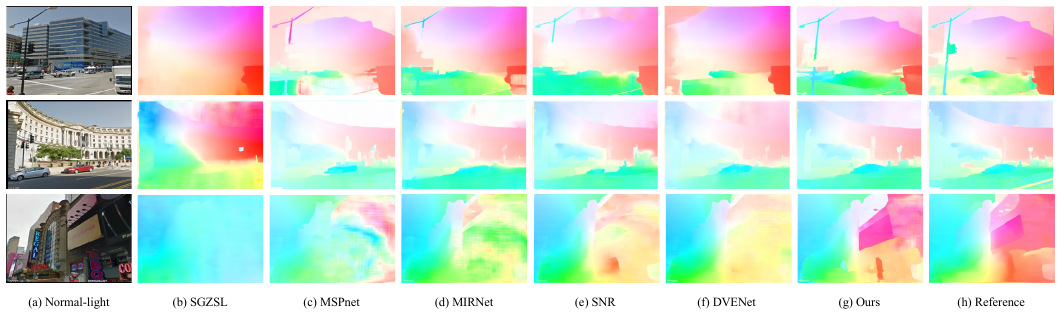}
\caption{Visual comparison of optical flow estimation results by different low-light enhancement methods.}
\label{fig-opticalflow-response}
\end{figure*}

\begin{table*}[t]
  \centering
  \caption{Multi-view Consistency comparison in terms of AB, MABD, and E$_{warp}$. The values highlighted with bold font and underlined indicate ranking first and second place, respectively.}
  \label{tab-consistency}
  \scalebox{0.9}{
    \begin{tabular}{c|c|c|c|c|c|c|c}
    \toprule
    Method & NPE \cite{wang2013naturalness} & LIME \cite{guo2016lime} & SGZSL \cite{zheng2022semantic} & L3Fnet \cite{lamba2020harnessing} & MSPnet \cite{wang2023multi} & RetinexNet \cite{wei2018deep} & MBLLEN \cite{lv2018mbllen} \\
    \midrule
    AB $\downarrow$ & 22.52 & 12.22 & 26.55 & 11.79 & 18.14 & 17.88 & 31.03 \\
    MABD($\times10^{-2}$) $\downarrow$ & 0.5825 & 0.4347 & 0.3516 & 0.5464 & 0.6180 & 1.0906 & 0.2446 \\
    E$_{warp}$($\times10^{-2}$) $\downarrow$ & 2.064 & 3.257 & 2.741 & 1.552 & 2.257 & 1.795 & 2.238 \\
    \midrule
    Method & ZeroDCE \cite{guo2020zero} & DSLR \cite{lim2020dslr}  & MIRNet \cite{zamir2020learning} & SGM \cite{yang2021sparse} & SNR \cite{xu2022snr} & DVENet \cite{huang2022low} & Ours \\
    \midrule
    AB $\downarrow$ & 27.40 & 8.13 & 7.91 & 9.34  & 9.43 & \underline{6.85} & \textbf{6.75}  \\
    MABD($\times10^{-2}$) $\downarrow$ & 0.5192 & 0.1907 & 0.2874 & 0.3054 & \textbf{0.1448} & 0.3690 & \underline{0.1700}  \\
    E$_{warp}$($\times10^{-2}$) $\downarrow$ & 2.528 & 1.674 & \underline{1.506} & 1.625 & 1.630 & 1.536 & \textbf{1.339} \\
    \bottomrule
    \end{tabular}}
\end{table*}

\textbf{Qualitative Results.} We perform the visual comparisons on the MVLT dataset and real-world scenes to evaluate the performance of different methods qualitatively. Fig.~\ref{fig-syn} shows the enhanced results of diverse methods on our synthesized MVLT dataset. As can be observed, there exists visible noise artifacts and undesirable color deviation for SGZSL \cite{zheng2022semantic} and MSPnet \cite{wang2023multi}, respectively. For LLFlow\cite{wang2022low}, we could notice lightness attenuation and over-smoothing texture destruction in the restored images, which lead to the weak naturalness. Compared to the MIRNet \cite{zamir2020learning} and DVENet \cite{huang2022low}, our method consistently achieves better enhancement with more appealing visual quality. We further present qualitative comparisons on the real-world scenes in Fig.~\ref{fig-real}, which are captured by Canon EOS R6 with different ISO settings. As can be seen, SGZSL \cite{zheng2022semantic} tends to overexpose the low-light inputs with intensive noise. In general, our method achieves better visual enhancement than the state-of-the-art methods on noise removal, detail preservation, and color consistency.

\textbf{Consistency Analysis.} Following the recent low-light video enhancement works \cite{jiang2019learning, lv2018mbllen}, we adopt the Average Brightness variance (AB) and Mean Absolute Brightness Difference (MABD) to evaluate the brightness consistency. To verify the content consistency, the Warping Error (E$_{warp}$) \cite{lai2018learning} among multi-views is calculated based on the optical flow estimation\cite{teed2020raft}. Herein, smaller values of AB, MABD, and E$_{warp}$ indicate better multi-view consistency. The results are shown in Table~\ref{tab-consistency}. As can be observed, our proposed method achieves a competitive consistency enhancement and gains the best AB and E$_{warp}$ values. Though our method achieves the second-best MABD result, the value is only 0.252$\times$10$^{-3}$ more than SNR \cite{xu2022snr}. In addition to the quantitative comparisons, we also provide the qualitative visualization in Fig.~\ref{fig-opticalflow-response}. The Reference optical flow estimated by the normal-light images is adopted for reference. We can observe that SGZSL \cite{zheng2022semantic} fails to accurately capture the structures and edges of pixel motion in adjacent views. Other methods either generate inaccurate predictions in local regions \cite{wang2023multi, huang2022low}, or struggle to estimate refined optical flow \cite{zamir2020learning, xu2022snr}. In contrast, our proposed method achieves more promising multi-view consistency between different viewpoints, approaching the quality of the ground truth as closely as possible.

\begin{table*}
\begin{minipage}[t]{0.5\textwidth}
\centering
\caption{Ablation study of network component settings in our RCNet, including intra-view enhancement (Intra-view EN) and inter-view alignment \& fusion (Inter-view AF) within each recurrent unit. The best result is highlighted in bold.}
\begin{tabular}{cc|c|c} % |c
    \toprule
    \multicolumn{2}{c|}{Network Setting} & \multicolumn{2}{c}{Quality Metric} \\
    \midrule
    Intra-view EN & Inter-view AF & PSNR$\uparrow$ & SSIM$\uparrow$ \\ % & LOE$\downarrow$
    \midrule
    \XSolidBrush & \XSolidBrush & 22.95 & 0.8615 \\ % & 000.0
    \Checkmark   & \XSolidBrush & 24.55 & 0.8715 \\ % & 000.0
    \XSolidBrush & \Checkmark   & 24.86 & 0.8794 \\ % & 000.0
    \Checkmark   & \Checkmark   & \textbf{26.21} & \textbf{0.8834} \\ % & \textbf{000.0}
    \bottomrule
\end{tabular}
\label{tab2-ablation-component}
\end{minipage}
\begin{minipage}[t]{0.5\textwidth}
\centering
\caption{Ablation study of network interaction settings in our RCNet, including interactive connections from Enhancement to Alignment (E2A) and from Alignment to Enhancement (A2E). The best result is highlighted in bold.} 
\begin{tabular}{cc|c|c} % |c
    \toprule
    \multicolumn{2}{c|}{Network Interaction} & \multicolumn{2}{c}{Quality Metric} \\
    \midrule
    E2A & A2E & PSNR$\uparrow$ & SSIM$\uparrow$ \\ % & LOE$\downarrow$
    \midrule
    \XSolidBrush & \XSolidBrush & 26.21 & 0.8834 \\ % & 000.0
    \Checkmark   & \XSolidBrush & 26.36 & 0.8828 \\ % & 000.0
    \XSolidBrush & \Checkmark   & 26.42 & \textbf{0.8845} \\ % & 000.0
    \Checkmark   & \Checkmark   & \textbf{26.45} & 0.8844 \\ % & \textbf{000.0}
    \bottomrule
\end{tabular}
\label{tab3-ablation-interaction}
\end{minipage}
\end{table*}

\begin{figure*}[b]
\centering
\includegraphics[width=\linewidth]{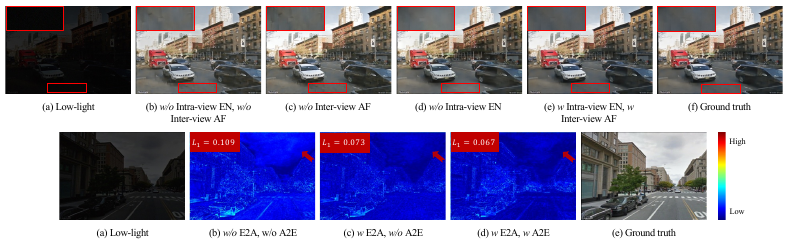}
\caption{Visual quality comparisons of the effectiveness of Intra-view EN and Inter-view AF (in the top row), as well as two different interactions E2A and A2E (in the bottom row) in our RCNet.}
\label{fig-settings}
\end{figure*}

\subsection{Ablation Study}
In this subsection, we conduct ablation studies to investigate the effectiveness of our RCNet with several network variants, including different network component settings and interactions, as well as the number of recurrent units. 
% For fair comparison, we take the results of nine epochs as an average to make a reliable validation with diverse network setting.
% \begin{table}[t]
%     \centering
%     \caption{Ablation study of network component settings in our RCNet, including intra-view enhancement (Intra-view EN) and inter-view alignment \& fusion (Inter-view AF) within each recurrent unit. The best result is highlighted in bold.}
%     \begin{tabular}{cc|c|c} % |c
%     \toprule
%     \multicolumn{2}{c|}{Network Setting} & \multicolumn{2}{c}{Quality Metric} \\
%     \midrule
%     Intra-view EN & Inter-view AF & PSNR$\uparrow$ & SSIM$\uparrow$ \\ % & LOE$\downarrow$
%     \midrule
%     \XSolidBrush & \XSolidBrush & 22.95 & 0.8615 \\ % & 000.0
%     \Checkmark   & \XSolidBrush & 24.55 & 0.8715 \\ % & 000.0
%     \XSolidBrush & \Checkmark   & 24.86 & 0.8794 \\ % & 000.0
%     \Checkmark   & \Checkmark   & \textbf{26.21} & \textbf{0.8834} \\ % & \textbf{000.0}
%     \bottomrule
%     \end{tabular}
%     \label{tab2-ablation-component}
% \end{table}

\textbf{Investigation of Network Component Settings.} As the core components of our RCNet, Intra-view EN and Inter-view AF are able to extract the discriminative intra-view features and perform feature alignment between the \textit{primary view} and each \textit{auxiliary view}, respectively. To validate the effectiveness of these two components, we explore four different network settings within our RCNet, and the comparative results of RCNet and its three variants are listed in Table~\ref{tab2-ablation-component}. As can be observed, when we first remove both Intra-view EN and Inter-view AF in our RCNet, the average values of PSNR and SSIM suffer severe decreases as compared with the RCNet. Due to the absence of two core components, the model often cannot recover the texture details via multi-view collaboration and is prone to produce unexpected artifacts in the enhanced results. It is worth noting that this model could be improved significantly when applying the Intra-view EN or Inter-view AF individually. Moreover, our RCNet can achieve the best performance because of the combination of Intra-view EN and Inter-view AF, which could be further validated by a visual quality comparison provided in the top row of Fig.~\ref{fig-settings}.

% \begin{table}[t]
%     \centering
%     \caption{Ablation study of network interaction settings in our RCNet, including interactive connections from Enhancement to Alignment (E2A) and from Alignment to Enhancement (A2E). The best result is highlighted in bold.}
%     \begin{tabular}{cc|c|c} % |c
%     \toprule
%     \multicolumn{2}{c|}{Network Interaction} & \multicolumn{2}{c}{Quality Metric} \\
%     \midrule
%     E2A & A2E & PSNR$\uparrow$ & SSIM$\uparrow$ \\ % & LOE$\downarrow$
%     \midrule
%     \XSolidBrush & \XSolidBrush & 26.21 & 0.8834 \\ % & 000.0
%     \Checkmark   & \XSolidBrush & 26.36 & 0.8828 \\ % & 000.0
%     \XSolidBrush & \Checkmark   & 26.42 & \textbf{0.8845} \\ % & 000.0
%     \Checkmark   & \Checkmark   & \textbf{26.45} & 0.8844 \\ % & \textbf{000.0}
%     \bottomrule
%     \end{tabular}
%     \label{tab3-ablation-interaction}
% \end{table}

\textbf{Effectiveness of Interactions between Enhancement and Alignment.} Table~\ref{tab3-ablation-interaction} shows the ablation investigation on the effects of network interaction from Enhancement to Alignment and from Alignment to Enhancement. It can be analyzed from the table that when removing these two network interactions E2A and A2E, the RCNet suffers from an undesirable quality degradation in terms of PSNR value with 0.24dB and SSIM value with 0.001. This is because the proposed RCNet without the E2A connection cannot perform inter-view alignment from auxiliary views adaptively depending on the enhancement quality, while the RCNet removing the A2E connection makes it difficult to enhance the intra-view images for missing similar feature propagation without these two network connections between enhancement and alignment stage. When applying the E2A and A2E interactive connection alone, we can see that the quality of the enhanced result is improved in PSNR and SSIM of different degrees. To this end, our method RCNet equipped with two network interactions E2A and A2E can achieve the best enhancement result in terms of PSNR and SSIM. In the bottom row of Fig.~\ref{fig-settings}, we report the $L_{{\rm 1}}$ distance between the enhanced results and normal-light image when the E2A or A2E is ablated. As can be seen, the $L_{{\rm 1}}$ distance tends to be larger when ablating the interaction E2A or A2E. In comparison, the enhanced result obtained from our RCNet (\textit{w} E2A, \textit{w} A2E) exhibits a more promising result, revealing the efficacy of incorporating both the E2A and A2E interactions in enhancing multi-view low-light images.

\textbf{Investigation of Number of Recurrent Units.} We further investigate the enhancement performance of RCNet with a diverse number of recurrent unit ReEAF. As shown in Table~\ref{tab4-ablation-number}, the scheme of recurrent feature enhancement, alignment and fusion can significantly improve the enhancement quality of resulted images. More specifically, the quality gaps between ReEAF-1 and ReEAF-2 are large, which demonstrates that the cascaded ReEAFs can aggregate inter-view contextual details from the previous unit and guide the next unit to restore the \textit{primary view} effectively. Besides, when compared to the SSIM gains, the PSNR metric achieves considerable improvements. One of the most possible reasons is that the second ReEAF performs the attentive spatial enhancement on the most similar aligned features propagated from the first ReEAF, which can bring accurate pixel fidelity for the performance boost. In addition, though the quality gaps between ReEAF-2 and ReEAF-3 are marginal, we still could observe that the third ReEAF brings consistent improvements by the feature refinement from multiple views.
\begin{table}[t]
    \centering
    \caption{Investigation of the number of recurrent unit ReEAF in our RCNet. Note that ReEAF-\textit{N} indicates \textit{N} ReEAFs adopted in total. The best result is highlighted in bold.}
    \begin{tabular}{c|c|c|c}
    \toprule
    Quality Measure & ReEAF-1 & ReEAF-2 & ReEAF-3 \\
    \midrule
    PSNR $\uparrow$ & 25.41 & 26.34 & \textbf{26.45} \\
    SSIM $\uparrow$ & 0.8770 & 0.8826 & \textbf{0.8844} \\
    % PSNR $\uparrow$ & 25.41 & 26.34 & \textbf{26.70} \\
    % SSIM $\uparrow$ & 0.8770 & 0.8826 & \textbf{0.8848} \\
    % LOE$\downarrow$ & 0.000 & 0.000 & \textbf{0.000} \\
    \bottomrule
    \end{tabular}
    \label{tab4-ablation-number}
\end{table}

\begin{figure*}[t]
\centering
\includegraphics[width=\linewidth]{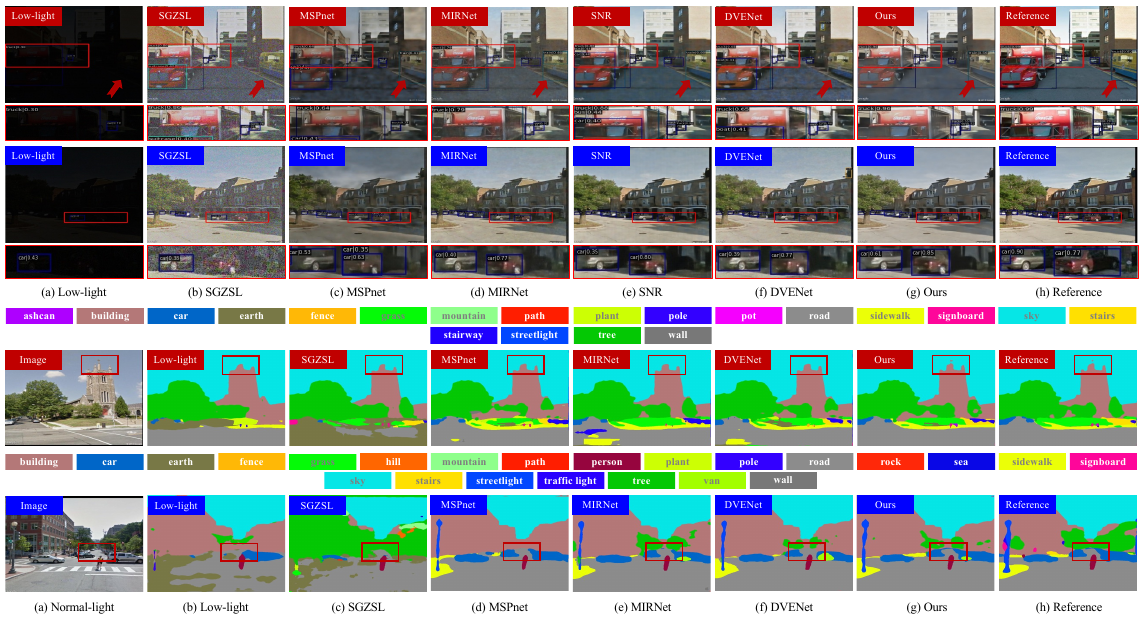}
\caption{Visual comparisons of two object detection algorithms Faster R-CNN \cite{ren2015faster} and RetinaNet \cite{lin2017focal} (in the top two rows), as well as two semantic segmentation algorithms PSPNet \cite{zhao2017pyramid} and DeepLabv3+ \cite{chen2018encoder} (in the bottom two rows) among different low-light enhancement methods.}
\label{fig-tasks}
\end{figure*}

\subsection{Application for High-level Tasks}
In order to validate the improvements of our proposed method on outdoor recognition tasks, we adopt two popular object detection algorithms Faster R-CNN \cite{ren2015faster} and RetinaNet \cite{lin2017focal}, as well as two semantic segmentation algorithms PSPNet \cite{zhao2017pyramid} and DeepLabv3+ \cite{chen2018encoder} to detect/segment the enhanced results generated by different low-light enhancement methods. 
%Furthermore, we use the pretrained models of two detectors trained on the COCO \cite{lin2014microsoft} and two segmentors trained on the ADE20K \cite{zhou2017scene} within the same backbone ResNet50 \cite{he2016deep}. 
Here we provide more descriptions regarding the performance improvements of our method on the object detection and semantic segmentation tasks, respectively.

\textit{Object Detection}. Object detection aims to recognize bounding boxes and classes of the objects in the input image. Herein, the multi-view low-light images are first enhanced by different low-light enhancement methods, and then the object detectors \cite{ren2015faster, lin2017focal} are performed for the performance comparison. From the first two rows in Fig.~\ref{fig-tasks}, we observe that low-light images present a dilemma: some objects are detected with low precision, and in some cases, they cannot be recognized regardless of the detector used. However, the low-light enhancement methods can alleviate this dilemma to some certain extent. More specifically, the enhanced results of SGZSL \cite{zheng2022semantic} exhibit the capability to detect either trucks (in the first row) or cars (in the second row) with high precision. However, the inadequate noise removal in this method results in the generation of several inaccurate bounding boxes for a single object, thereby impacting the overall accuracy and reliability of the detection results. Compared to existing low-light enhancement methods, including the multi-frame method SGZSL \cite{zheng2022semantic}, multi-view methods MSPnet \cite{wang2023multi} and DVENet \cite{huang2022low}, and the single image-based methods MIRNet \cite{zamir2020learning} and SNR \cite{xu2022snr}, our proposed multi-view low-light image enhancement method obtains the competitive detection precision consistently when different detection algorithms utilized.

\textit{Semantic Segmentation}. In the last two rows of Fig.~\ref{fig-tasks}, we present visual quality comparisons of the segmentation results when different enhancement models are utilized. Two different segmentation algorithms \cite{zhao2017pyramid, chen2018encoder} are performed on each individual scene, respectively. Intuitively, our proposed method yields more promising segmentation results compared to recent state-of-the-art methods, as can improve the accuracy of true category labels while reducing the occurrence of false labels.
For example, false classification of the `earth' category rather than the `road' can be observed in the segmentation results of low-light image and SGZSL \cite{zheng2022semantic}, which are not presented in our segmentation result. Moreover, our method achieves a competitive prediction on the true pixels of `tree' and `car', as depicted using the red rectangle in the last row.

\begin{table*}[t]
  \centering
  \caption{Model complexity comparison on parameter size (Param), FLOPs, and inference time (Time). All the models are evaluated with the input image size set as 256$\times$256. Note that Ours$_{{\rm N}}$ means there are N ReEAFs adopted in RCNet.}
  \label{tab-complexity}
  \scalebox{0.9}{
    \begin{tabular}{c|c|c|c|c|c|c|c|c|c}
    \toprule
    Method & ZeroDCE \cite{guo2020zero} & DSLR \cite{lim2020dslr} & SGZSL \cite{zheng2022semantic} & MSPnet \cite{wang2023multi} & MIRNet \cite{zamir2020learning} & SNR \cite{xu2022snr} & Ours$_{1}$ & Ours$_{2}$ & Ours$_{3}$ \\
    \midrule
    Param(M) & 0.08  & 14.93 & 0.01 & 1.18 & 31.79  & 39.12 & 2.23 & 3.99 & 5.75  \\
    FLOPs(G) & 10.38 & 11.75 & 0.09 & 605.30 & 1632.31 & 47.92 & 1283.56 & 2433.83 & 3584.10  \\
    Time(s)  & 0.187 & 0.158 & 0.165 & 0.133 & 0.169 & 0.08 & 0.281 & 0.582 & 0.882  \\
    \midrule
    PSNR(dB)$\uparrow$ & 15.71 & 23.34 & 16.58 & 19.90 & 25.05 & 25.72 & 25.41 & 26.34 & 26.45  \\
    LOE$\downarrow$    & 757.0 & 246.6 & 547.7 & 349.1 & 164.9 & 124.3 & 170.5 & 136.3 & 124.8  \\
    \bottomrule
    \end{tabular}}
\end{table*}%

\subsection{Discussion for Model Complexity}
For a more comprehensive comparison, we further evaluate our model against recent works in terms of parameter size, FLOPs, and inference time. In particular, we tested all models using the same image size (256$\times$256), and the inference time was evaluated on a Nvidia GeForce RTX 3090. The results are presented in Table~\ref{tab-complexity}. From the table, we can observe that the parameter size of our method is nearly one-eighth of the second-best method (SNR \cite{xu2022snr}). However, the FLOPs and inference time of our method are not the best. This is primarily due to the top-K patches searching process during cross-view alignment. It is worth noting that the model complexity can be reduced by adjusting the value of K to a smaller one. We further explore the model complexity when using different number of recurrent unit ReEAF in our RCNet. As can be seen, the enhancement performance achieves considerable improvements with increasing parameter size and inference time when integrates more recurrent units. Nevertheless, we believe that the trade-off in model complexity is meaningful when our main objective is to achieve the best enhancement results. Therefore, we set the N=3 (\textit{i.e.}, Ours$_{3}$) as our final method.

\section{Conclusion} \label{sec6-conclusion}
In this paper, we make the first attempt to investigate multi-view low-light image enhancement. First, we construct a new dataset called Multi-View Low-light Triplets (MVLT), including 1,860 pairs of triple images with large illumination ranges and random noise distribution. Each triplet is equipped with three different viewpoints towards the same scene. Second, we propose a deep multi-view enhancement framework based on the Recurrent Collaborative Network (RCNet). In order to benefit from similar feature correspondence across different views, we design the recurrent feature enhancement, alignment and fusion (ReEAF) module, in which intra-view feature enhancement (Intra-view EN) followed by inter-view feature alignment and fusion (Inter-view AF) is performed to model the intra-view and inter-view feature propagation sequentially via multi-view collaboration. In addition, we develop two different interactions E2A and A2E between Intra-view EN and Inter-view AF, which utilize the quality-aware feature weighting for similar patches and attentive spatial sampling, respectively. Experimental results demonstrate that our RCNet significantly outperforms recent state-of-the-art methods.

% \section*{Acknowledgments}
% This should be a simple paragraph before the References to thank those individuals and institutions who have supported your work on this article.

\bibliographystyle{IEEEtran} % mybstfile
\bibliography{IEEEfull,IEEEpre} % mybibfile

\newpage
% If you have an EPS/PDF photo (graphicx package needed), extra braces are needed around the contents of the optional argument to biography to prevent the LaTeX parser from getting confused when it sees the complicated $\backslash${\tt{includegraphics}} command within an optional argument. (You can create your own custom macro containing the $\backslash${\tt{includegraphics}} command to make things simpler here.)


% \begin{IEEEbiography}[{\includegraphics[width=1in,height=1.25in,clip,keepaspectratio]{figures/fig1.png}}]{Michael Shell}
% \end{IEEEbiography}



\vfill

\end{document}